\documentclass[conference]{IEEEtran}
\IEEEoverridecommandlockouts
\usepackage{cite}
\usepackage{amsmath,amssymb,amsfonts}
\usepackage{algorithmic}
\usepackage{graphicx}
\usepackage{textcomp}
\usepackage{xcolor}
\def\BibTeX{{\rm B\kern-.05em{\sc i\kern-.025em b}\kern-.08em
    T\kern-.1667em\lower.7ex\hbox{E}\kern-.125emX}}
\usepackage{mathtools}
\usepackage{multirow}
\usepackage{float}
\usepackage{siunitx}
\usepackage{listings}
\usepackage{svg}
\usepackage[linesnumbered,ruled,vlined]{algorithm2e}
\usepackage{hyperref}

\begin{document}

\title{A Connected Component Labelling algorithm for multi-pixel per clock cycle video stream\\
\thanks{The work presented in this paper was supported by the National Science Centre project no. 2016/23/D/ST6/01389 entitled ``The development of computing resources organisation in latest generation of heterogeneous reconfigurable devices enabling real-time processing of UHD/4K video stream''.}
}

\author{\IEEEauthorblockN{Marcin Kowalczyk}
\IEEEauthorblockA{\textit{Department of Automatic Control and Robotics} \\
\textit{AGH University of Science and Technology}\\
Krakow, Poland \\
kowalczyk@agh.edu.pl}
\and
\IEEEauthorblockN{Tomasz Kryjak \textit{Senior Member IEEE}}
\IEEEauthorblockA{\textit{Department of Automatic Control and Robotics} \\
\textit{AGH University of Science and Technology}\\
Krakow, Poland \\
tomasz.kryjak@agh.edu.pl}
}


\maketitle


\begin{abstract}
This work describes the~hardware implementation of a~connected component labelling (CCL) module in reprogammable logic.
The main novelty of the design is the ``full'', i.e. without any simplifications, support of a~4~pixel per clock format (4~ppc) and real-time processing of a~4K/UltraHD video stream (3840 x 2160 pixels) at 60 frames per second. 
To achieve this, a~special labelling method was designed and a~functionality that stops the input data stream in order to process pixel groups which require writing more than one merger into the equivalence table.
The proposed module was verified in simulation and in hardware on the \textit{Xilinx Zynq Ultrascale+ MPSoC} chip on the \textit{ZCU104} evaluation board.
\end{abstract}

\begin{IEEEkeywords}
FPGA, Zynq UltraScale+ MPSoC, 4K, UHD, real-time video processing, connected component labelling (CCL)
\end{IEEEkeywords}

\section{Introduction}

Connected  component  labelling  is  among  the  most widely used image processing algorithms.
Its purpose is to assign the same label to all pixels that belong to one object (a connected group of pixels).
The input to the algorithm is an image after binarisation, containing only the values 0 (the pixel belongs to the background) or 1 (the pixel belongs to the foreground object). 
On the other hand, the output is an image of the same size, whose pixels are the values of the assigned labels.
Two pixels are assumed to belong to the same object if there is a~path between them that contains only the foreground pixels.
The operation of the algorithm depends, among others, on the type of neighbourhood considered. 
Typically an 8-element neighbourhood is used, but a~4-element neighbourhood can also be applied.


The field of computer vision is constantly evolving.
Much of this progress is related to the development of new data processing algorithms, but the vision sensors themselves are also getting better.
Over the years, they offered higher and higher refresh rates and higher resolutions.
Recently, Ultra High Definition (UHD or 4K -- \(3840 \times 2160\)) resolution is getting closer to being common.
Even mid-range smartphones are already able to record an image in this resolution while maintaining 60 frames per second.
The use of high resolution sensors allows not only to improve the quality of the displayed image, but also to increase the efficiency of vision algorithms --  provide a~more accurate image of an object that is further away from the camera, and consequently increases the chances of the correct object's recognition.
This is important, e.g. in advanced driver assistance systems (ADAS), surveillance systems or issues related to autonomous vehicles (cars or UAVs).



However, increasing the resolution has a~huge impact on the amount of data needed to be processed. 
A video stream with UltraHD resolution at 60 frames per second results in a~data flow of \(1424 \si{MB/s}\).
Processing this amount of data in real time requires considerable computing power. 
One platform that can meet this requirement is the \textit{Xilinx Zynq Ultrascale+ MPSoC} heterogeneous chip, which includes an ARM processor system and reprogrammable logic (FPGA -- Field Programmable Gate Array).


According to the authors' knowledge, the presented solution is the only module tested in hardware, which, by processing the video stream transmitting four pixels in one clock cycle, allows for labelling the video stream with UltraHD resolution and 60 frames per second in real-time.
The main contributions of this paper are:
\begin{itemize}
    \item design of a~hardware architecture that performs the connected component labelling algorithm (CCL) of an UltraHD resolution video stream transmitting four pixels in one clock cycle at 60 frames per second (to the best knowledge of the authors, this is the first such architecture evaluated in hardware),
    \item utilisation of the AXI4-Steam capability to temporarily stop the data stream in order to process the most complicated pixel combinations,
    \item verification of the proposed architecture on the \textit{ZCU104} evaluation board equipped with the \textit{Zynq Ultrascale+ MPSoC} chip by \textit{Xilinx}.
\end{itemize}




This paper has been divided into the following parts:
Section \ref{sec:previous} discussed research related to CCL and CCA algorithms and their FPGA implementation.
Then, in Section \ref{sec:algorytm} the~proposed connected component labelling algorithm is described.
Finally, the designed architecture implementing the algorithm is described in Section \ref{sec:architektura}.
Evaluation of the architecture is presented in Section \ref{sec:ewaluacja}.
The paper ends with a~conclusion and possible further research directions.

\section{Previous work}
\label{sec:previous}

Connected component labelling is a~fundamental algorithm in the field of computer vision. Because of that, many articles dealing with the problem of its hardware implementation have been published.

We presented a~comprehensive state-of-the-art review in our previous paper \cite{kowalczyk2021real}. Below we discuss only papers that were not analysed in that article and briefly summarise the review.

In the work \cite{klaiber2019single} an architecture that allows to process several pixels of an image simultaneously is presented. 
For this purpose, the input image is divided into several vertical segments. 
Each segment is transferred to a~separate data processing module and is treated as a~separate image. 
At the same time, an additional module collects information on relations between objects that span several segments. 
Relationships between such objects are saved to the global segment graph (GSG), which is created while processing the segments. 
This graph allows to save information about connections between different segments and to solve the resulting connections. 
The architecture was described in the VHDL hardware description language and tested on the FPGA \textit{Xilinx Virtex 6 VLX240T-2} chip. 
The declared bandwidth of the architecture should be sufficient to process data with UltraHD resolution in real-time. 
However, no information about the operation of the system for the real-time vision data stream is provided.


The article \cite{ling2017fpga} presents a~hardware architecture of the CCL algorithm without the use of an equivalence table. 
For this purpose, a~structure is created mapping the connections between the given labels. 
The authors claim that the presented architecture is able to process images with a~resolution \(256 \times 256\) in time \(0.44\si{ms}\), with a~clock frequency of \(150 \si{MHz}\). 
This translates to 2,273 frames per second.


The authors of the article \cite{bailey2019zig} provide a~solution that allows to overcome the problem of resolving conflicts at the end of an image line by using zig-zag scanning instead of the typical raster scan. 
This allows the chain of mergers to be resolved while processing the next line of the frame. 
The proposed architecture can process 1-bit binary pixel per clock cycle. 
The authors report that the maximum frequency is equal to \(180\si{MHz}\), which is not enough to process UltraHD at 60 frames per second video stream in real-time.

In the paper \cite{ciarach2019real} a~hardware implementation in reconfigurable of a~single-pass connected component labelling and connected component analysis module is presented.
The proposed design supports real-time processing of a~UltraHD/4K video stream at 60 frames per second.
The architecture is tested in a~skin colour area segmentation problem.
The design processes a~video stream in 4 pixels per clock cycle format, but a~major simplification was done in order to achieve this.
The proposed solution consists in connecting two adjacent (binary) pixels using the "OR" operator.
This approach greatly simplifies the problem, but comes down to reducing the horizontal resolution of the input video stream.

An architecture realising the CCL algorithm was also presented in the work \cite{kowalczyk2021real}, which is a~continuation and extension of the paper \cite{ciarach2019real}.
In this article an analysis on possible solutions for the UltraHD video stream labelling problem was conducted.
An architecture for processing a~video stream in 2 pixels per clock cycle format was also presented.
In this case no simplifications were necessary.
Drawback of this approach is the doubled operating frequency (\(300\si{MHz}\) was used), which results in higher energy dissipation.

Summing up the review, it should be noted that it covers a~period of 25 years (1995-2020).
There was a~dynamic development of technology during that time -- both in vision sensors and FPGA devices.
For example, in the work \cite{rachakonda_1995} from 1995, 9 Xilinx XC4010 chips were used and the image was processed with a~resolution of \(512 \times 512\) at 30 fps (clock frequency \(10 \si{MHz}\)).
In the paper \cite{spagnolo_2020} from 2020, a~single Zynq SoC was used for a~stream with a~resolution of \(2K \times 2K\) and over 30 fps (clock frequency around \(100 \si{MHz}\)).
In our previous work \cite{ciarach2019real} we have proposed a~system capable of processing a~simplified video stream in 4 pixels per clock cycle format.
Then, in \cite{kowalczyk2021real} we designed an architecture processing a~video stream in 2 pixels per clock cycle format without any simplification.
However, until now, no architecture capable of processing a~video stream in 4 pixels per clock cycle was presented.

It should be noted that the progress, understood as the possibility of processing a~stream with higher resolution and fps, is firstly related to the use of newer generations of computing platforms and peripherals (like HDMI 2.0 input/output modules).
However, a~direct analysis of this phenomenon and an attempt to compare (reduce to the ``common denominator'') solutions in terms of the use of different computing platforms is not simple and was not the aim of this article.
The second factor enabling the aforementioned progress is the use of various algorithmic solutions, which consequently translate into the hardware architecture of the CCL module.
The mentioned solutions can be divided into the following categories:
\begin{itemize}
    \item two-pass -- \cite{rachakonda_1995} (1995), \cite{spagnolo_2020} (2020),
    \item two-pass with pixel ''series'' analysis -- \cite{appiah_2010} (2010),
    \item single-pass -- \cite{bailey_2008} (2008), \cite{klaiber_2016} (2016), \cite{ling2017fpga} (2017), \cite{spagnolo_2019} (2019), \cite{klaiber2019single} (2019), \cite{bailey2019zig} (2019), \cite{ciarach2019real} (2019), \cite{kowalczyk2021real} (2021),
    \item single-pass with shift register -- \cite{jeong_2016} (2016),
    \item single-pass with advanced pixel ``series'' analysis -- \cite{zhao_2017} (2017), \cite{tang_2018} (2018),
    \item single-pass with post processing on an ARM core -- \cite{spagnolo_2018} (2018).
\end{itemize}

In each of the above-mentioned publications, the authors introduce some algorithmic improvements, which, apart from the use of newer generation equipment, allow to achieve better video stream processing parameters.
In addition, recently (2016-2020) authors have been looking for algorithmic improvements through the use of e.g. shift register and pixel series analysis.
The aim of these works is, among others, the optimisation of resource utilisation, or the desire to eliminate calculations during horizontal and vertical blanking.
The other aim of the proposed improvements is to increase the throughput of the architecture, which enables to process more data in the same amount of time, which is necessary to process data from sensors of higher resolution or more frames per second.



\section{The Proposed CCL Algorithm}
\label{sec:algorytm}


For UltraHD resolution and 60 frames per second, the data stream frequency is about \(600\si{MHz}\).
This is much more than the maximum frequency even for modern FPGAs.
A possible solution to this issue is data transfer parallelisation -- in one clock cycle, 2 or 4 image pixels can be processed, thereby reducing the stream frequency correspondingly two or four times.
A CCL module supporting such a parallelised video stream was the subject of earlier works -- \cite{ciarach2019real}, \cite{spagnolo_2020} and \cite{kowalczyk2021real}.
However, so far it has not been possible to perform a~``full'' connected component labelling for 4~pixels per clock cycle, without reducing the image resolution horizontally (thus applying some simplifications). Because of this, the proposed two-pass connected component labelling algorithm was designed.

\begin{figure}[!t]
	\centering
	\includegraphics[width=3in]{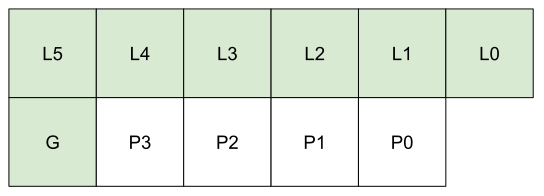}
	\caption{Neighbourhood for the processed group of four pixels.}
	\label{fig:sasiedztwo}
\end{figure}

For a~4~ppc video stream, in each clock cycle, the neighbourhood of 11 pixels should be analysed. 
This is presented in Figure \ref{fig:sasiedztwo}.


Labels marked L and G are assumed to be known. 
In a~given clock cycle, however, the labels for four pixels from P3 to P0 have to be determined. . 
For each pixel, a~new label is derived based on the neighbourhood from the previous row and the label on the left. 
This means that the labelling must be sequential. 
There is no way to parallelise the labelling process.
Then, in the next clock cycle, the label given to pixel P0 will be saved as pixel label G.



\begin{figure}[!t]
	\centering
	\includegraphics[width=3in]{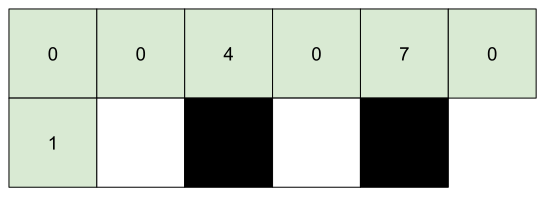}
	\caption{An example neighbourhood that requires the resolution of two conflicts.}
	\label{fig:przekodowanienadania1}
\end{figure}

Moreover, attention should be paid to the fact that in the described situation up to two conflicts (two different labels in the same neighbourhood of the currently considered pixel)  may occur in one clock cycle, which will have to be resolved by modifying the equivalence table. 
An example of such a~situation is shown in Figure \ref{fig:przekodowanienadania1}.

\begin{figure}[!t]
	\centering
	\includegraphics[width=3in]{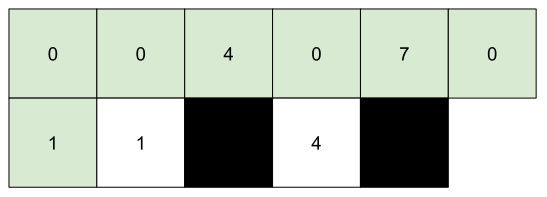}
	\caption{Labels assigned after processing subsequent pixels in the group.}
	\label{fig:przekodowanienadania2}
\end{figure}

In the presented case, label 1 will be determined for pixel P3 and a conflict between labels 1 and 4 will be reported. 
For pixel P1, label 4 will be assigned and a conflict between labels 4 and 7 will be reported.
It should be noted that when a merger is detected, a higher label is always assigned to the lower one in the equivalence table and the lower label is assigned to the processed element.
After processing the subsequent pixels in the group, the labels shown in Figure \ref{fig:przekodowanienadania2} will be assigned.


In addition to labelling, two conflicts will be designated: \(+4 \rightarrow 1\) and \(7 \rightarrow 4\). 
It is necessary to analyse the detected mergers so as not to create redundant connections in the equivalence table. 
These conflicts should point to a~single label. Analysis is required only when two mergers are detected for a pixel group.
In result, the following connections are determined: \(4 \rightarrow 1\) and \(7 \rightarrow 1\).

\begin{figure}[!t]
	\centering
	\includegraphics[width=3in]{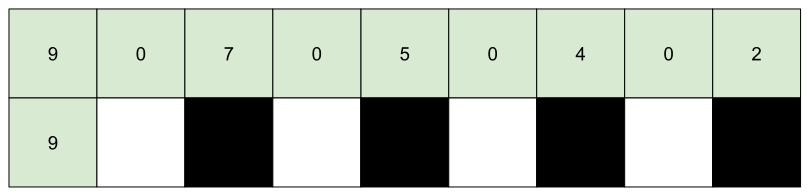}
	\caption{Example situation of mergers chain.}
	\label{fig:lancuch1}
\end{figure}

For each conflict, a~decision is also made whether or not a~given conflict should be entered into the chain of mergers.
Chain of mergers is a situation when consecutive labels would be merged to lower and lower labels. An example of such a situation is presented in Figure \ref{fig:lancuch1}.

In the presented situation, the following conflicts will be detected: \(9 \rightarrow 7\), \(7 \rightarrow 5\), \(5 \rightarrow 4\) and \(4 \rightarrow 2\). The merger chain is used to avoid creating complicated connections in the equivalence table. They would require a frequent and computationally demanding search for the root of the graph. Instead, such mergers are written into the stack, which is resolved between consecutive lines of the image. For the considered case, as a result of the module's operation, label 2 would be entered in the equivalence table for cells 9, 7, 5 and 4.

\begin{figure}[!t]
	\centering
	\includegraphics[width=3in]{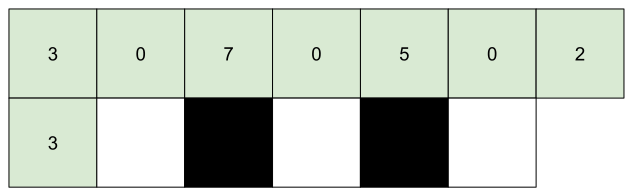}
	\caption{Example situation of additional merger chain.}
	\label{fig:lancuch2}
\end{figure}

However, there is an additional situation when it is necessary to use a merger chain. This situation is shown in Figure \ref{fig:lancuch2}.

The first four pixels without a label form one group. In this situation, only two conflicts should be added to the merger chain by default.
However, this will cause the indirect connection -- \(5 \rightarrow 3 \rightarrow 2\) -- in the equivalence table. Such a connection would require a much more complicated search for the corresponding label.
In order to prevent indirect connections, it was decided to enter both conflicts of a given group into the merger chain in such a case.
In this way, label 5 will be directly connected to label 2.
Pseudocode of the merger analysis process is presented in the Algorithm \ref{alg:merger_analysis}.

\begin{algorithm}[!t]
 \KwData{merge1 - first merger\newline
         merge2 - second merger}
 \KwResult{merge1 - first modified merger\newline
           merge2 - second modified merger\newline
           stack1 - first merger chain flag\newline
           stack2 - second merger chain flag}
    \uIf{\(merge1(1) = merge2(1)\)}{
        \eIf{\(merge1(2) > merge2(2)\)}{
            \(merge1(1) \gets merge1(2)\)\;
            \(merge1(2) \gets merge2(2)\)\;}{
            \(merge2(1) \gets merge2(2)\)\;
            \(merge2(2) \gets merge1(2)\)\;}
        \(stack1 \gets 1\)\;
        \(stack2 \gets 1\)\;}
    \uElseIf{\(merge1(1) = merge2(2)\)}{
        \(merge2(2) \gets merge1(2)\)\;
        \(stack1 \gets 0\)\;
        \(stack2 \gets 0\)\;}
    \uElseIf{\(merge1(2) = merge2(1)\)}{
        \(merge1(2) \gets merge2(2)\)\;
        \(stack1 \gets 1\)\;
        \(stack2 \gets 1\)\;}
    \Else{
        \(stack1 \gets 1\)\;
        \(stack2 \gets 0\)\;}
 \caption{Mergers analysis process}
 \label{alg:merger_analysis}
\end{algorithm}

\begin{figure}[!t]
	\centering
	\includegraphics[width=3in]{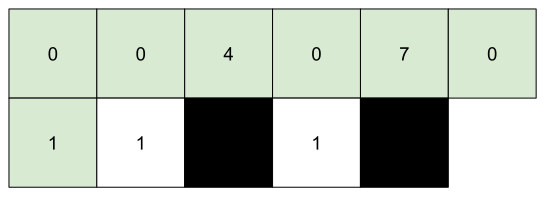}
	\caption{Final labels after recoding -- all condisterd pixels have the value '1'}
	\label{fig:przekodowanienadania3}
\end{figure}

The assigned labels are then recoded on the basis of the determined mergers. 
The final labels are shown in Figure \ref{fig:przekodowanienadania3}. 
The labels are then entered into the resulting image.

Between consecutive lines of the image, the conflicts that were entered into the stack containing the chains of mergers are processed and then written into the equivalence table.
After processing the entire image frame, a~final recoding is necessary. 
Polega ono na odczytaniu etykiety z każdej komórki tablicy przekodowań. Dla każdej odzytanej etykiety sprawdzany jest jej odpowiednik w tej samej tablicy. Odpowiednik ten jest wpisywany do pierwotnej komórki.
It consists in reading the label of each cell in the equivalence table. For each label, its equivalent in the same table is checked. This equivalent is entered into the original cell. The process is briefly presented in the Algorithm \ref{alg:final_recoding}.

\begin{algorithm}[!t]
 \KwData{eqTable - original equivalence table}
 \KwResult{eqTable - modified equivalence table}
 \ForEach{\(label \in eqTable\)}{
    \(label \gets eqTable(label)\)\;
    }
 \caption{Final recoding algorithm}
 \label{alg:final_recoding}
\end{algorithm}

Then, the image created from previously assigned labels is recoded based on the final equivalence table.


\section{The Proposed Hardware Architecture}
\label{sec:architektura}

\begin{figure*}[ht]
	\centering
	\includegraphics[width=5in]{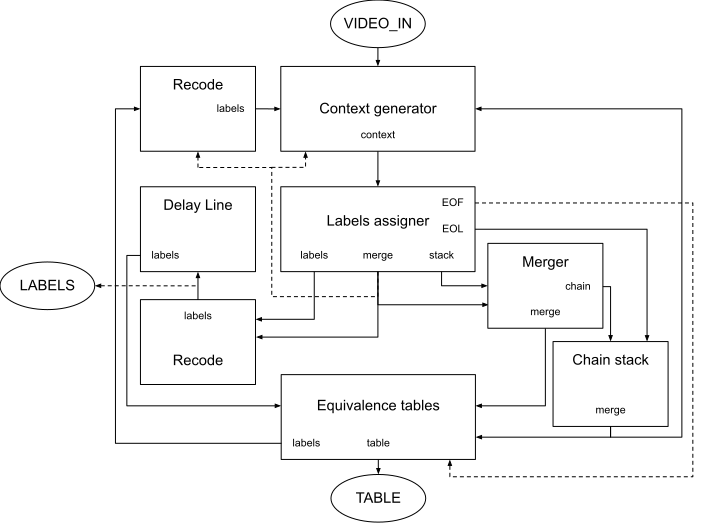}
	\caption{Diagram of the proposed hardware architecture.}
	\label{fig:architektura}
\end{figure*}

The hardware architecture realising the connected component labelling algorithm described in Section \ref{sec:algorytm} was designed taking into account the specificity of the problem and the authors' experience acquired during the research described in papers \cite{ciarach2019real} and \cite{kowalczyk2021real}. 
In the second of these works, an analysis of potential solutions to the 4~ppc problem was presented, which was the basis for creating the introduced algorithm. 
Its schematic is shown in Figure \ref{fig:architektura}.

The binary input stream \texttt{VIDEO\_IN} uses the \textit{AXI4-Stream} interface. 
It contains the signals \texttt{tvalid}, \texttt{tdata}, \texttt{tuser}, \texttt{tready}, and \texttt{tlast}.
The \texttt{LABELS} output stream contains the same signals.
The output \texttt{TABLE} includes the signals \texttt{addr} (equivalence table address), \texttt{data} (equivalence table data), and \texttt{valid}.
The parameters of the designed architecture are the number of bits for labels (and thus the maximum number of labels) and the resolution of the processed video stream.

\subsection*{Context generator}
\label{context_generator}

The binary data stream enters the \texttt{Context generator} module.
In addition to the input data stream, the module also receives a~stream of recoded labels from the \texttt{Delay Line} module, conflicts determined by the \texttt{Label assigner} module, and conflicts read at the end of each line from the stack containing the chains of mergers.


The module determines the position of the processed pixel group in the current frame. 
If the group belongs to the first line of the image (pixels L5 to L0 in the figure \ref{fig:sasiedztwo}), then the neighbourhood labels belonging to the previous line of the image are considered to belong to the background and zeros are inserted in their place in the context. 
If the group contains the left edge of the image, 0 is put in place of L5. In this case, the label of pixel G is changed to 0 inside the \texttt{Label assigner} module. 
On the other hand, if the group contains the right edge of the image, the L0 pixel label is changed to 0. In the rest of the image, the module does its default task and passes the label values from the previous image line.

\begin{figure*}[ht]
	\centering
	\includegraphics[width=6.8in]{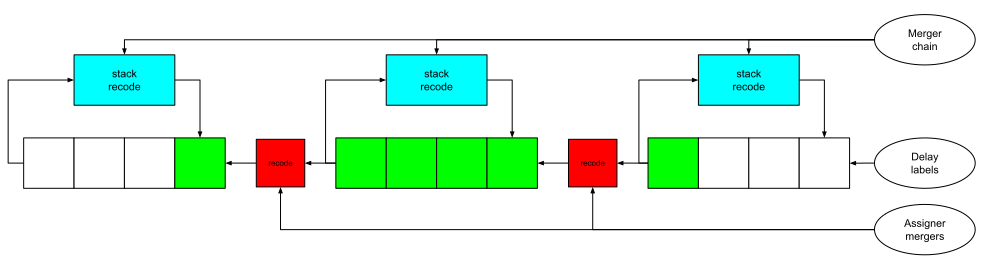}
	\caption{Context generation diagram.}
	\label{fig:context_generation}
\end{figure*}

This module also checks if a~group of input pixels is valid. 
This situation corresponds to the high state on the \texttt{tready} and \texttt{tvalid} signals of the input stream. 
If the group is valid, it is necessary to shift and recode the groups in context.
Recoding of the context is also necessary when processing the merger chain stack between consecutive lines of the frame.
In this case, however, new data is not received from the \texttt{VIDEO\_IN} data stream, so the groups in the context are not shifted.
After recoding, they must be put in the same place of the context. 
This is shown in Figure \ref{fig:context_generation}. 

The red modules only work when the pixel group of the \texttt{VIDEO\_IN} interface is valid. They recode the labels in the context based on the mergers received from \texttt{Label assigner} module. Labels after recoding are written into the next group in the context.
The blue modules work when the content of the merger chain stack is analysed (between consecutive lines of the processed frame). They recode the labels in the context based on the mergers received from \texttt{Chain stack} module. These modules write the recoded labels back into the same place in the context.
The green pixels belong to the context and are passed to the \texttt{Label assigner} module.

\subsection*{Label assigner}
\label{label_assigner}

The entire \texttt{Label assigner} module only works for valid pixel groups. 
It consists of several elements. 
First, the input context and the pixel on the left are recoded based on the conflicts determined for the previous valid pixel group. 
At the beginning of the frame (determined on the basis of the \texttt{tuser} signal), the global counter of the assigned labels is reset. 
This counter is used to label pixels that have only background pixels in their vicinity.
There is a possibility to design a label reuse module that would allow to recover labels that were merged with a different one and are no longer used.
Such a module will be designed in the future version of the architecture.


Then there are four smaller modules that process the neighbourhood of successive pixels. 
A~given module only works when the corresponding pixel belongs to the foreground object. 
Each module creates a~binary vector based on the neighbourhood. 
The created vector controls a~multiplexer. 
On its basis, appropriate actions are performed that are needed to determine a~new label and detect a~possible conflict. 
There must also be a~test, whether a~given conflict should be entered into the chain of mergers. 
The designated label is passed to the next module. 
The last module also updates the global counter of the assigned labels.


After all labels have been assigned, it is necessary to analyse the identified conflicts. 
The operation of the module depends on the number of designated conflicts. 
If no conflicts have occurred, no action is required. 
If one conflict is detected, the appropriate data should be simply passed to the output of the module. 
For two conflicts, considerably more operations are required.
It is caused by the necessity to conduct the analysis that was presented in the Algorithm \ref{alg:merger_analysis}.
In the latter case, the \texttt{pause} flag is set high for one clock cycle. 
It means that data processing must be suspended. 
This is due to the fact that only one cell of the equivalence table can be changed in one clock cycle. 
By pausing the data pipeline, it is possible to include both mergers in the equivalence table.


At the end of the module, the label on the right side of the group is assigned to the G~context pixel (fig. \ref{fig:sasiedztwo}) if the group is not the last in the line (signal \texttt{tlast}). Otherwise, pixel G~is assigned a~label 0 (representing a~pixel belonging to the background).
Control signals are also generated for the data stream \texttt{LABELS} (fig. \ref{fig:architektura}).
All operations described above have to be performed in one clock cycle. This is caused by the fact that the detected mergers are utilised in the next clock cycle to recode the context labels.

\subsection*{Merger}
\label{merger}


The \texttt{Merger} module is responsible for entering the appropriate data into the memory that stores the equivalence table. 
As the input, it accepts the information about conflicts from the \texttt{Labels assigner} module and the end of line signal of the label stream. 
On the output, it gives addresses and write data to the appropriate memories (equivalence table and stack of merger chains).
The module works only when the number of input conflicts is different from 0 and the data is valid (signal \texttt{tvalid} of the input \texttt{merge} stream).


At the end of the image line, the stack address is moved to the beginning of the memory (address 0). The address points to the top of the stack that is inside of the \texttt{Chain stack} module (described later in this section).


The module is designed so that in one clock cycle it can receive up to two conflicts at the input. 
In this case, the module processes them for two clock cycles (this is possible thanks to stopping the pipeline in the \texttt{Labels assigner} module -- flag \texttt{pause}). 
The larger label of the considered conflict is passed as the equivalence table address to the module output, and the smaller label is passed as the data.
If a~given conflict is marked as a~chain of mergers, the address pointing to the top of the stack is incremented. 
Both labels are written into the stack at that address.

\subsection*{Chain stack}
\label{chain_stack}

The main element of the \texttt{Chain stack} module is a~BRAM, which stores the stack containing chains of mergers. 
The input to the module is the address of writing to the memory, data that should be saved at the given address and the signal indicating the end of the image line. 
The output of the module is the address and data used for updating the equivalence table.


During processing a~line, the data received from the \texttt{Merger} (containing labels of the conflict) is written to memory, and after the \texttt{EOL} (end of line) input signal, the data is read from the memory. 
The larger of the read labels is transferred as the write address to the equivalence table, and the smaller one as the data to be written under this address.

\subsection*{Equivalence tables}
\label{equivalence_tables}


The \texttt{Equivalence tables} module inputs are the address and data stream from the \texttt{Merger} module, the same type of stream from the \texttt{Chain stack} module, and the stream of labels from the \texttt{Delay Line} module.
The outputs of the module are the stream of recoded labels from the delay line and the data stream of the equivalence table at the end of each video frame.


The main elements of the module are two memory banks, which are swapped between successive frames (double buffering). 
This is due to the need to reinitialise the equivalence tables in such a~way that at the beginning of a~new frame, each cell of the table points to itself, e.g. the memory location with the address 100 must contain the value 100. 
Each bank contains five BRAM memory modules. 
Each memory contains the same equivalence table.


Four memory modules are required to recode each of the four pixels in the group received from the \texttt{Delay Line} module in one clock cycle. 
An additional memory module is used to recode the data read from the memory that stores the merger chains. 
The same write address, data, and \texttt{Write Enable} signal are passed to each memory module. 
This ensures data consistency between them.


Each memory bank can be in one of three states. 
The first state represents a~typical bank operation, where conflicts are written into the array and the input labels from the \texttt{Delay Line} are recoded.


The second state involves the final recoding of the table. 
The bank enters this state after the frame processing is completed. 
The data from the first memory module of a~given bank is read  sequentially and transferred as an address to the next module. 
The data read from the second module is passed to the \texttt{TABLE} data stream.
This operations have to be done after frame processing is finished, as doing this operation during processing of the frame would increase the latency of writing mergers into the equivalence table. In the proposed solution, the data is written in one clock cycle. 
Additional recoding of the label would delay solving equivalences by the read latency of the memory. 
In effect, many more recoding operations would have to be performed.
This is the final equivalence table for the previous image frame.


The third state is responsible for the initialisation of the considered memory bank. 
The bank enters this state after the final recoding is completed. 
The addresses of each equivalence table cell are generated sequentially, and the same address is passed as data to the cells. 
Once the initialisation is complete, the bank waits for the frame processing to complete. 
Then it goes back to the first state.


Between successive lines of the processed frame, it is necessary to include the data from the stack of merger chains to the equivalence table.
However, the data from the chain must first be recoded. 
The address (larger of the labels) does not require recoding. 
For this purpose, the data received from the \texttt{Chain stack} module is first passed as an address to the fifth memory module. 
The read label from this memory is transferred as write data to all modules of a~given bank. 
The write address is an appropriately delayed signal obtained from the \texttt{Chain stack} module.


The pixel groups obtained from the \texttt{Delay Line} module are transferred to the appropriate memory bank. 
Each pixel label is passed as a~read address to one of the four memory modules. 
The group recoded in this way is transferred to the module output. 
Note that the read latency of the memory modules is equal to 1.

\subsection*{Delay Line}
\label{delay_line}


The \texttt{Delay Line} module accepts the stream of recoded labels from the \texttt{Labels assigner} module as input. 
It outputs the same labels, but delays them so that they can be used for context generation when processing the next line of the image. 
The module's operation is based on BRAM memory, which performs the function of a~circular buffer. 
Its size is equal to the number of pixel groups in one line of the image reduced by the recoding latency, context size, \texttt{Labels assigner} latency and memory read latency.


The module only works when the pixel group of the \texttt{VIDEO\_IN} interface is valid. 
For each valid pixel group, the internal counter is incremented. 
However, if the value of the counter is equal to the size of the buffer minus 1, it is reset to 0. 
This counter is the address of the currently processed item in the circular buffer. 
The labels from this address are read and sent to the output, and the input labels are written in the same place (the BRAM is configured in the \textit{read first} mode).

\subsection*{Recode}
\label{recode}


There are two \texttt{Recode} modules in the architecture. 
The first one is responsible for recoding the labels coming from the \texttt{Label assigner} module, and the second one for recoding the labels coming from the \texttt{Equivalence tables} module.
The inputs of the module are the label stream and the conflict stream. 
The output is a~stream of recoded labels.



These recodings are necessary due to the latency of the modules. 
While recoding labels from the \texttt{Delay Line} module, not all conflicts have been written to the recoding table yet, because of the write and read latency of the BRAM memory.
Therefore, conflicts not written into the equivalence table yet must be used to recode the output labels from the \texttt{Equivalence tables} module.


Input to the \texttt{Delay Line} module is also recoded based on the found conflicts. 
The output of this module is assigned to the \texttt{LABELS} output interface.

\section{Evaluation}
\label{sec:ewaluacja}

The presented architecture is described in the hardware description language \textit{SystemVerilog}. Modules were tested separately in \textit{Vivado 2018.2} simulator. The complete system was also tested in the simulation. Tests were conducted for short input vectors and for whole images. These simulations contributed to the detection of many errors at the hardware architecture design stage. Full compliance of the simulation results with the software model has been achieved.

Later, the architecture was tested on the \textit{Xilinx Zynq Ultrascale+ MPSoC} chip on the \textit{ZCU104} evaluation board. 
\textit{Vivado 2018.2} was used for implementation of the architecture. 
The input data was streamed from a~graphics card of a~PC to provide appropriate test patterns (in order to compare results).


The bit width of the labels has been set to 10, which in effect allows to assign 1023 labels. The implementation was also done for 15 bit width (32767 labels), but no tests were conducted for this case.
The resolution of the processed video stream was \(3830 \times 2160\) at 60 frames per second. 
The frequency of the processing pipeline was set to \(133.3 \si{MHz}\). This frequency is sufficient to process the UltraHD resolution video stream at 60 frames per second.
The maximum clock frequency of the created system is equal to \(153 \si{MHz}\). The maximum frequency depends, inter alia, on the bit width of the labels. 
Power consumption of the Zynq chip is equal to \(4.67 \si{W}\), of which \(1.894 \si{W}\) is consumed by the reprogrammable logic.
The data was taken from the \textit{Vivado} power estimation.


Table \ref{tab:zasoby} shows the resource utilisation from the \textit{Place Design} utilisation report of the module with the parameters listed above.

\begin{table}[ht]
	\centering
	\caption{Resource utilisation}
	\begin{tabular}{| c | c | c | c |}
		\hline 
		{Resource} & {Pass-through} & {CCL} & {System}     \\	
		\hline
		LUT      & 37397 (16.23\%) & 1588 (0.69\%) & 41782 (18.13\%) \\	
		\hline
		LUTRAM   & 3233 (3.18\%)   & 0 (0.00\%)   & 3709 (3.64\%) \\
		\hline
		FF       & 43369 (9.41\%)  & 714 (0.15\%)  & 48366 (10.50\%)  \\
		\hline
		BRAM     & 6 (1.92\%)      & 7.5 (2.40\%)   &  35 (11.22\%)  \\
		\hline
		DSP      & 3 (0.17\%)      & 0 (0.00\%)    &  3 (0.17\%)   \\
		\hline
		BUFG     & 26 (4.78\%)     & 1 (0.18\%)    &  27 (4.96\%)   \\
		\hline
	\end{tabular}
	\label{tab:zasoby}
\end{table}
The module was tested with a~threshold module in order to binarize the input data stream before the poposed CCL module. A cumbersome test patterns were passed as an input to the HDMI input of the \textit{ZCU104} evaluation board. These test patterns required all modules to work as intended in order to get a correct result.
The outputs of the evaluated architecture were connected to an \textit{Integrated Logic Analyzer}, which allows to look at the connected signals during operation of the system. 
The equivalence table was compared with the reference software model of the algorithm, which was implemented in the \textit{MATLAB} computing environment.
The results were consistent.
A few lines of the \texttt{LABEL} data stream were also compared with the corresponding lines of the model result. 
The same labels were assigned in both cases.


\section{Conclusions}
\label{sec:podsumowanie}
In the paper, we proposed a~hardware architecture for the connected component labelling algorithm. 
It allows to process 4 pixels of the input stream (4~ppc format) in one clock cycle without any simplification.
The design is capable of working with UltraHD/4K video stream at 60 frames per second in real-time.
A key issue is the functionality that allows to temporarily stop the input data stream. 
It enables to process pixel groups, which require writing two conflicts into the equivalence table.
The designed solution is utilising a~small amount of computing resources (max. \(2.5\%\) of the used chip for the BRAM memory).

The program model of the algorithm in \textit{MATLAB} computing environment can be found on the GitHub:\newline 
\url{https://github.com/vision-agh/CCL}.



According to the authors' knowledge, the presented solution is the only module tested in hardware, which, by processing the video stream transmitting four pixels in one clock cycle, allows for labelling the video stream with UltraHD resolution and 60 frames per second in real-time.

The concept of processing a video stream with 8K resolution also seems worth analysing. In this case, it may be necessary to process 8 pixels in one clock cycle with twice the frequency. The most troublesome in this situation seems to be the \texttt{Label assigner} module, which has to determine the label values for consecutive pixels of one group in one clock cycle. Increasing the number of pixels in this module can have a very negative effect on the maximum operating frequency. However, thorough tests must be carried out to unambiguously answer this question.

The architecture will be further developed. 
We plan to design an additional module responsible for the calculation of the objects' parameters (bounding box, area, centroid). 
An additional module allowing to reuse the merged labels would be also very beneficial for the proposed solution.

\end{document}